# Causal Discovery of Flight Service Process Based on Event Sequence

Zhiwei Xing,[1,2] Lin Zhang,[1,2] Huan Xia,[2] Qian Luo,[*1,2] and Zhao-xin Chen,[2]

[1]School of Electronic information and Automation, Civil Aviation University of China, Tianjin, 300300, China.
[2] The Second Research Institute of Civil Aviation Administration of China, Chengdu, 610041, China.

Correspondence should be addressed to luoqian@caacetc.com

## Abstract

The development of the civil aviation industry has continuously increased the requirements for the efficiency of airport ground support services. In the existing ground support research, there has not yet been a process model that directly obtains support from the ground support log to study the causal relationship between service nodes and flight delays. Most ground support studies mainly use machine learning methods to predict flight delays, and the flight support model they are based on is an ideal model. The study did not conduct an in-depth study of the causal mechanism behind the ground support link and did not reveal the true cause of flight delays. Therefore, there is a certain deviation in the prediction of flight delays by machine learning, and there is a certain deviation between the ideal model based on the research and the actual service process. Therefore, it is of practical significance to obtain the process model from the guarantee log and analyze its causality. However, the existing process causal factor discovery methods only do certain research when the assumption of causal sufficiency is established and does not consider the existence of latent variables. Therefore, this article proposes a framework to realize the discovery of process causal factors without assuming causal sufficiency. The optimized fuzzy mining process model is used as the service benchmark model, and the local causal discovery algorithm is used to discover the causal factors. Under this framework, this paper proposes a new Markov blanket discovery algorithm that does not assume causal sufficiency to discover causal factors and uses benchmark data sets for testing. Finally, the actual flight service data is used.

## 1. Introduction

The 2019 Civil Aviation Industry Development Statistics Bulletin[1] shows that compared with 2018, the total civil aviation transportation turnover in 2019 has increased by 7.2%. The increase in the total airport transportation turnover requires the improvement of airport flight service efficiency. Flight service includes various activities, such as opening the cabin door, cleaning, refueling, etc. The flight service process is highly dependent, and the delay of a single node will affect the delay of subsequent operations, resulting in delays in the launch of flights. These attributes add additional complexity to the assurance operation. For example, to add aviation fuel for a departing flight, we need to know the type of aircraft of this flight, the position of the aircraft, the planned departure time of the aircraft, and other decision-making information, in order to determine when, where, and how much aviation fuel the tanker will carry to complete the flight. It is a complicated process in itself. However, its own attributes, such as the fuel capacity and integrity of the scheduled aviation fuel vehicle, will also affect whether the aviation fuel addition can be completed normally. The attributes of the service equipment itself make the work of aviation fuel addition is more complicated. In view of the highly dependent characteristics and complexity of the flight service process mentioned above, process managers usually do not know which operation nodes have problems during flight service operations that will ultimately affect the overall performance of flight service, Such as flight departure delays, the duration of refueling truck operations, etc. Therefore, when an airport is experiencing delays in the launch of departing flights, it is difficult for airport managers to find the service nodes that directly cause the delay and take targeted improvement measures to the problem nodes.

Although some scholars[2][3][4][5] have established flight service process models with methods such as critical path restoration, colored Petri nets, and Service model tools for the flight service process, and conducted in-depth studies on the scheduling of ground service resources, the flight service model in the above studies It is driven by a manually drawn model. This is an ideal model, which is deviated from the actual support business process model. The deviation is mainly reflected in the manually drawn ideal model. The node importance and topological sorting of each support node are only based on expert knowledge. As the managers of each airport are different, and their specific service plans have their own characteristics, the artificially drawn aircraft service model is not completely applicable to the flight service process of a specific airport. At the same time, the research on flight launch delays focuses on predicting the length of the delay, while the prediction method based on statistical correlation emphasizes the correlation between influencing factors and results rather than causality, so it is impossible to analyze which links ultimately lead to the flight launch delay, And due to the existence of unobservable variables, there is a certain deviation in the prediction results.

At present, flight support operations are supported by airport information systems. These systems record the history of outbound flight support operations in the form of event sequences. Therefore, in order to analyze the performance of flight support, avoid the problem of deviations in artificially established process models. This paper uses process mining to dig out the actual process model of outbound flight service from the sequence of events. Process mining is a tool that transforms event data into business insights. It bridges the gap between traditional model-based process analysis (such as simulation and other business process management techniques) and data-centric analysis techniques (such as machine learning and data mining)[6].

Through process mining, we get a practical business process model, and on this basis, calculate the performance indicators of each link to describe the operation status of each link. By analyzing the causal relationship between the performance indicators of each link and the flight departure delay, we can determine which node's operation situation affects the flight departure and launch, and provide effective guidance for reducing flight delays and improving airport operation efficiency.

There has not yet been a phenomenon in the existing flight support research that causal factors in the business process based on actual service logs have been discovered. This paper proposes a framework, automatic discovery of causal factors in the presence of latent causal variables based on process mining (ACLP), based on the flight service process model mined by the process mining algorithm, and on the premise of relaxing the sufficiency of causality, using the score-based maximum ancestor graph Markov blanket The algorithm (SMMB) generates a local ancestor map of the flight service launch delay and use the direction between nodes extracted by the process model as a supplement to the local ancestry graph to adjust the direction of the edge, to realize the automatic discovery of cause and effect of the flight service business performance. Our proposed framework combines the process mining method with the causality discovery method under the existence of unobservable variables. When the business process of flight service is unknown, a flight service process model that fits the actual situation is established based on the sequence of events with process mining. Aiming at the problem of unobservable variables in actual scenarios, the newly proposed SMMB algorithm is used to determine the causal relationship between business process performance indicators, and the extraction of causal factors from event data to explaining business process performance is realized. The SMMB algorithm proposed in this paper is based on the score-based local directed acyclic graph (DAG) discovery algorithm, that is, the score-based local learning (SLL) algorithm, which is extended according to the characteristics of the Maximal ancestral graph (MAG) Markov blanket (MB). It is a topology-based method. The neighbor set and spouse set of the target variable is constructed by the method of scoring. Then according to the relevant definition and inference of the area set proposed by [16], the adjacent area set is determined, and the complete MB is searched. Through the advantages of scoring method in searching the neighborhood set and spouses set of the target variable, the SMMB algorithm has a better performance on the F-measure evaluation index than the constraint-based MAG MB algorithm proposed before. It provides new ideas for the automatic causal discovery of the flight service process under latent variables. This article is organized as follows: Section 2 discusses the background, Sections 3 and 4 respectively introduce the proposed framework and experimental results, and give some suggestions based on the experimental results. Section 5 draws conclusions and makes future work Outlook.

## 2. Related Work

### 2.1 Process model discoveries.

Whether defined and prescribed or implicit and temporary, business processes drive and support most of the functions and services in today's world's enterprises and management organizations. Based on the complexity of process control flow and the related concepts of process repeatability and predictability, the research of Di Ciccio et al. Di Ciccio et al.[8] divide business processes into the following three macro types: structured processes, semi-structured processes, Unstructured process. The structured process is characterized by a clearly defined, predictable, and repeatable sequence of activities and its input and output are pre-defined, while the semi-structured and unstructured process has no pre-defined and repeatable sequence of activities. The semi-structured process can outline the possible sequence of activities based on the case, determine the input of the required activities, and change the sequence of some activities through a specific situation's characteristics. The activities of the unstructured process are differently combined based on the specific instance. The sequence of activities becomes completely case-dependent when the level of process flexibility and unpredictability is increased.

Process mining is a method of analyzing actual business processes based on event logs generated by the system. The idea is to discover, monitor, and improve real business processes by extracting knowledge from event logs.

**Definition 1** (event [9]): An event is the instantiation of an activity in a business process, usually represented by a tuple $e=(a, caseID, t_{start}, t_{end}, d_1,...,d_m)$ , where $a$ represents the activity name attribute corresponding to the event, *caseID* represents the instance attribute where the event is located, and *eventID* represents the event *ID* attribute of the event, $t_{start}$ represents the start timestamp attribute of the event, $t_{end}$ represents the end timestamp attribute of the event, and $d_1,...,d_m(m \geq 0)$ represents other attribute values, where $\forall i \in [1,m], d_i \in D_i$ , $D_i$ represent the value range of each attribute. The event log L for a specific process model comprises a series of events in the process instance. The sequence of all events in the process instance in chronological order is the trajectory. A complete trajectory corresponds to one execution of the process. In terms of flow, all historical execution traces constitute the event log L.

The discovery of process models has always been a hot issue in process mining; that is, information about the original process model, organizational context, and execution attributes can be obtained from the execution log in the absence of a priori. Most process discovery algorithms usually apply a single algorithm to control flow steps[10], such as alpha algorithm[11], heuristic mining algorithm[12], multi-stage process mining algorithm[13], region-based mining algorithm[14]. The above algorithm is effective when applied to a structured business process. Still, when applied to a semi-structured or unstructured process, the model found by the above algorithm is really "spaghetti-like."These models describe every detail of unstructured behavior found in logs too finely. The reason for the problem lies in the assumptions that these process mining algorithms are based on. Assume as follows:

**Assumption 1[14]: All logs are reliable and trustworthy.**

**Assumption 2[14]: There exists an exact process which is reflected in the logs.**

These assumptions are completely reasonable in a structured and controlled environment, but they do not hold true in a less structured real environment. Therefore, the process model discovery algorithm based on the above assumptions will simulate the entire process completely, accurately, and meticulously. The results are often "spaghetti," and process managers cannot obtain effective information from the model.

In order to solve the above problems, Christian W et al. [15] proposed a process mining algorithm based on fuzzy theory. When dealing with unstructured problems, the fuzzy algorithm can distinguish whether the task is important or not and can remove unnecessary details. A more advanced view is abstracted, which focuses on discovering a more advanced mapping of behavior in the log rather than trying to discover the true process model.

2.2 Causal discovery algorithm.

Let $V=\{V_1,...,V_N\}$ contain N observed variables, P be a discrete joint probability distribution over V, and G represents DAG. we call the triple $<P,V,G>$ a Bayesian network(BN),if $<P,V,G>$ satisfies the Markov condition: each variable is independent of any subset of its nondescendant variables conditioned on its parents in G. The causal relationship between variables in BN can be represented by DAG G containing only directed edges (→).

**Definition 2** (Causal Sufficiency [16]): The observed variable set V is said to be causal sufficient if and only if any common cause of two or more variables in V is also in V.

Causal sufficiency considers that given a set of observed variables V, there is no latent common cause for V's subset of variables.

**Definition 3** (Faithfulness [17]): In a $BN<P,V,G>$, G is faithful to the probability distribution P over V if and only if every independence present in P is entailed by G and Markov conditions. P is faithful if and only if G is faithful to P.

The faithfulness assumption establishes a relationship between the probability distribution P and its underlying DAG G. We can use a conditional independence test instead of d separation to find all BN's dependencies or independence under this assumption. Under the assumption of satisfying the sufficiency of causality, the MB of the target variable in the DAG includes the parents, children, and spouses of the target variable T. Nowadays, the MB discovery algorithm for DAG has been relatively complete. It can be divided into topology-based methods and non-topological methods. The non-topological methods greedily test each variable and target by using the definition of Markov blanket, like the IAMB algorithm[18]. The topology-based approach aims to gradually search for the MB of the target node, such as Min-max Markov Blanket (MMMB)[19], using the topological characteristics of the MB. The article[20][21] introduce the same framework based on the topology method and conducts extensive experimental research to verify its superior performance in various applications.

Without assuming the sufficiency of causality, when the underlying data generated has potential common causes, MAG is proposed to represent the independent relationship between the observed variables. There is no need to mark the potential common causes in the structure clearly. A hybrid graph is a collection of nodes and edges, and its edges may be one-way edges (→) or bidirectional edges (↔). Suppose there is no directed ring (the presence of $V_i \to V_j$ and $V_i \leftarrow V_j$ at the same time) and almost directed ring (the presence of both $V_i \leftrightarrow V_j$ and $V_i \leftarrow V_j$) in the mixed graph, it is called an ancestor graph. On the path τ of the ancestry graph, if path τ contains $* \to V_i \leftarrow *$, then the non-endpoint variable $V_i$ is the colliding node in the ancestry graph. Otherwise, $V_i$ is a non-colliding node on τ. Every non-endpoint variable on the collision path from the target node T to Y in the MAG is a collision node. For example, $T \leftrightarrow V_1 \leftrightarrow V_2 \leftrightarrow V_3 \leftrightarrow Y$ is a collision path, and all $V_1$、$V_2$、$V_3$ is collision nodes.

**Definition 4** (v-structure [16]): In an ancestry graph, a triple $\{V_i, V_j, V_k\}$ is an unshielded triple if $V_i$ and $V_j$ are adjacent, and $V_j$ and $V_k$ are adjacent, but $V_j$ and $V_k$ are not adjacent. An unshielded triple $\{V_i, V_j, V_k\}$ is called a v-structure if $V_j$ is a collider on the path $\{V_i, V_j, V_k\}$, and the triple satisfies $\exists Z \subseteq V \setminus \{V_i, V_j, V_k\}$ such that $V_i \perp V_j \mid Z$ and $V_i \not\perp V_j \mid \{Z \cup V_j\}$ .In a v-structure $V_i \rightarrow V_j \leftarrow V_k$ , $V_k$ is the spouse node of $V_i$ .

**Definition 5** (m-connection, m-separation [22]): In the ancestor graph G=(E,V), given a set of nodes $Z, Z \subseteq V \setminus \{A, B\}$ , if it satisfies 1. The non-colliding nodes on the path p do not belong to Z. 2. Each colliding node on the path is the ancestor of a member of Z, so the path p between A and B is m-connection. If there is no m-connection path concerning Z between A and B, then A and B are m-separation.

**Definition 6** (Maximum Ancestor Graph [23]): For any two nonadjacent variables in an ancestor graph, if there is a set of variables m-separating them , the ancestor graph is maximal.

There are relatively few discovery algorithms for MAG MB. The article [16] proposed for the first time the constraint-based local causality discovery algorithm (M3B algorithm) of MB under the MAG framework that does not assume the sufficiency of causality, instead of learning the overall MAG and directly learning MB. This algorithm is a topology-based MB algorithm. The algorithm first finds the neighborhood set (parents and children) of the target node and uses a recursive search algorithm to recursively find the area set of a given target to complete the MB. The article [23] has proved that the constraint-based method is sensitive to error propagation, and there is no scoring method for the algorithm of MAG MB discovery. Therefore, the framework of the SLL algorithm mentioned is extended according to the characteristics of MAG MB. First, the neighbor set and spouse set of the target variable is constructed by the method of scoring. Then according to the relevant definition and inference of the area set proposed in the literature [7], the adjacent area set of the target node is determined, and the MB is finally completed.

This paper optimizes on the basis of the fuzzy process model and obtains the actual flight guarantee process model. Based on the actual business process, the performance indicators of the nodes are calculated to measure the operating status of each node, and finally, the SMMB local causality mining algorithm is used to find the root cause of the delay in flight launch when existing latent variables. This method provides new research ideas for discovering causal factors in process mining in the presence of latent variables. For the automatic discovery of causal factors of business performance, few scholars have done in-depth research. For example, literature [24][25] proposed a method for automatically discovering process performance bottlenecks and deviations based on event data, but it did not explore causality. Literature [26] proposed a method based on time series analysis to detect the causal relationship between business process characteristics and process performance indicators. However, the Granger causality test method adopted did not consider the existence of latent variables; that is, it believed that the actual data satisfies the assumption of causality sufficiency. Therefore, this article's contributions are as follows: (1) For the first time, it is proposed to realize the automatic discovery of causal factors of business performance under the premise of relaxing the assumption of causal adequacy. (2) The proposed local MAG discovery algorithm based on scoring has more advantages than M3B and RFCI algorithms.

# 3. Materials and Methods

This section proposes a framework to realize the automatic discovery of flight guarantee causality from the sequence of the events, as shown in Figure 1. The framework consists of two parts: (1) process model mining (2) construction of the local causal structure. The first part is the fuzzy mining algorithm, which mainly includes two stages. The first stage is the initialization stage, which establishes the initial process model through the flight guarantee event sequence. The second stage is the simplified stage, which mainly includes three parts, conflict resolution, edge filtering, Clustering, and abstraction. This article proposes solutions to the unary and N-ary conflicts in the initial model and optimizes the process model. The second part is the construction of local causal construction. This method is based on the SLL algorithm. It is mainly used to search for the neighboring nodes and spouse nodes of the target node and combines MAG MB based on the searched neighboring node-set and spouse set. The feature searches the area set of the target node, the parent set of the area set, etc., and then completes the complete Markov blanket's construction.

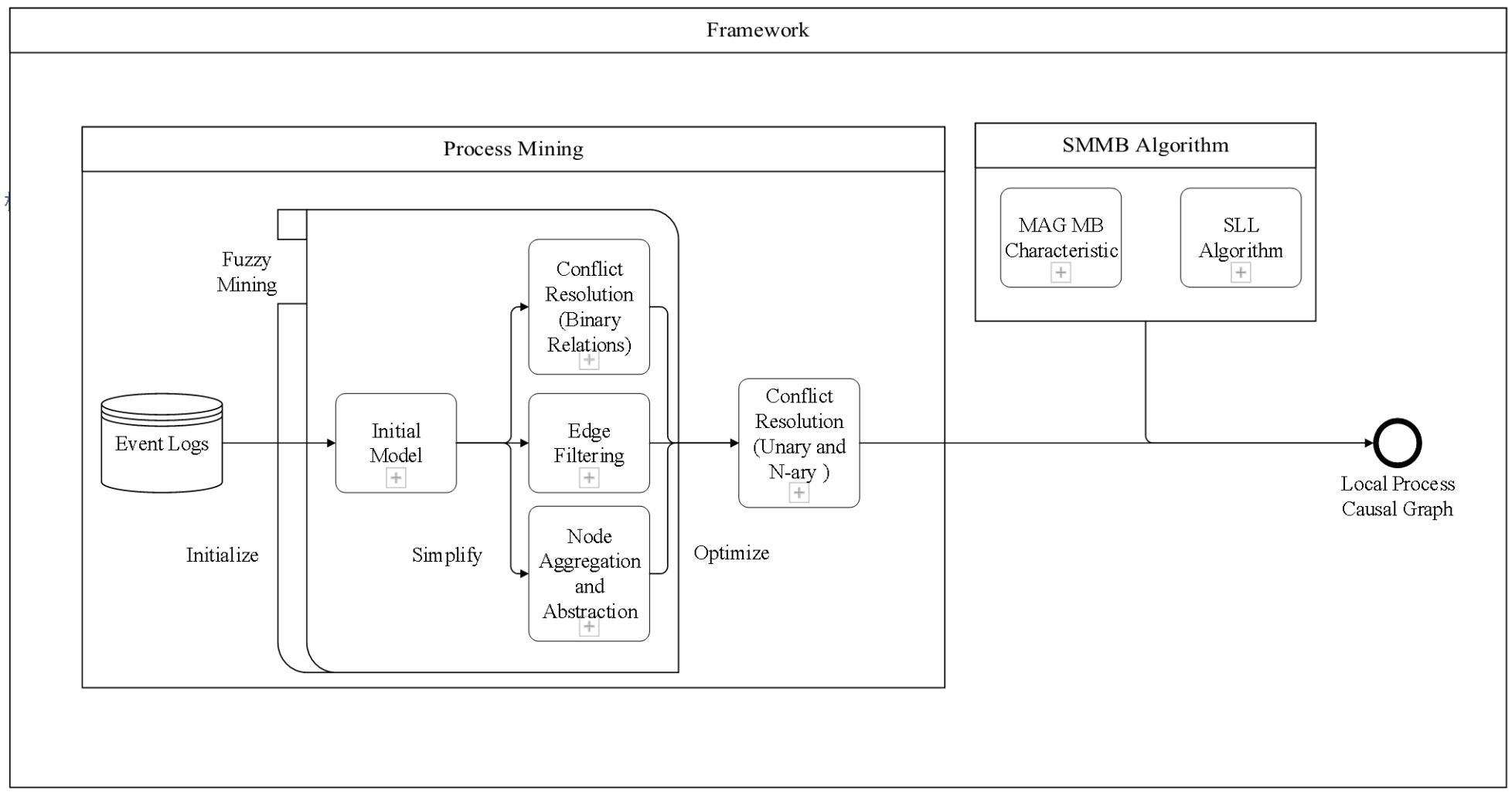

Figure 1: Automatic discovery framework of causal factors based on process mining

## 3.1 Process model mining

As mentioned earlier, the fuzzy mining algorithm is divided into two stages. The first stage is the initialization stage, which converts each observed event type into an activity node. The directed edges added to the model represent the ordering relationship between activities. The second stage is the simplification stage, divided into three steps: conflict resolution of binary relations, edge filtering, aggregation, and abstraction. As shown in Figure 2, the initialization model's possible conflict relationships include binary conflicts and N-ary conflicts, and unary conflicts. The conflict resolution part of the fuzzy mining algorithm only includes the solution of the binary conflict problem, which leads to the phenomenon of the N-ary cycle and self-circulation in the process model obtained by the fuzzy process mining algorithm. For the flight service process, self-circulation is possible in the process. Take the change of aircraft parking position as an example. When the aircraft enters the airport, due to the shortage of aircraft space resources, the aircraft needs to change its parking position several times. Therefore, in the event sequence, the change of aircraft parking position will appear multiple times in a row, and the model obtained by the process mining algorithm will have a self-circulation phenomenon. The existence of this phenomenon may be a pure exception. There are no binary and N-ary cycles in the actual operation process of flight service, and it operates in sequence over time. Aiming at the reason for this phenomenon and the solution of binary conflict of fuzzy mining algorithm, the solution of N-ary conflict and unary conflict is derived to optimize the process model of fuzzy mining.

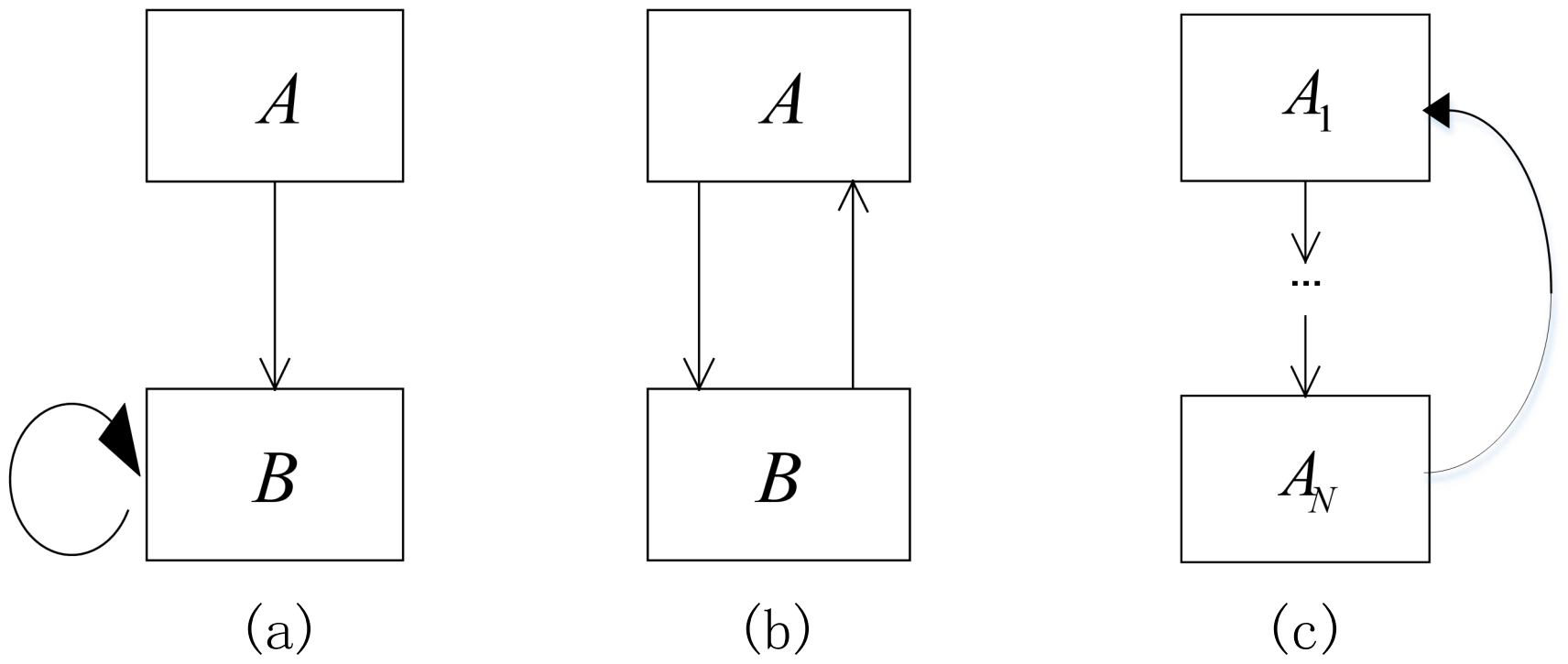

Figure 2: Conflicting relationships in the initial model. (a) Unary conflict.(b) binary conflict.(c) N-ary conflict.

In the fuzzy mining algorithm[27], the generation of binary conflicts is divided into three situations: binary loop, exception, and concurrency. If the relative importance $rel(A,B)$ and $rel(B,A)$ of two conflicting relationships exceed the retention threshold, then activities A and B form a binary cycle. If at least one conflicting relationship is lower than this threshold, determine the offset between the relative importance, $S(A,B) = rel(A,B) - rel(B,A)$ . If the offset value exceeds the ratio threshold, the less important relationships are deleted. Suppose at least one relationship wants to retain the threshold for importance, and the offset value is less than the ratio threshold. In that case, the relationship between A and B is a low and balanced relationship, implying that A and B are executed at the same time, so both edges are deleted at the same time. The formula for relative importance is as follows:

$$rel(A,B) = \frac{1}{2} \times \frac{sig(A,B)}{\sum_{X \in \eta} sig(A,X)} + \frac{1}{2} \times \frac{sig(A,B)}{\sum_{X \in \eta} sig(X,B)} \tag{1}$$

Where $\eta$ is the node-set of the process model, $sig : \eta \times \eta \to R_0^+$ is the priority relationship assigned to each pair of nodes $A, B \in \eta$ , and $rel : \eta \times \eta \to R_0^+$ is the relative importance between each pair of nodes A and B.

On the basis of the solution of the binary conflict, the solution of the N-ary conflict is derived. First, the relative importance of the binary relationship does not apply to the N-ary conflict, so it needs to be expanded to the relative importance of the N-ary relationship.

$$\begin{aligned} rel_i(A_i, A_{i+1}, ..., A_N, A_1, ...A_{i-1}) = & \frac{1}{2\ (N-1)} \times \frac{sig(A_i, A_{i+1})}{\sum_{X \in \eta} sig(A_i, X)} + \frac{1}{2\ (N-1)} \times \frac{sig(A_i, A_{i+1})}{\sum_{X \in \eta} sig(X, A_{i+1})} \\ & + \frac{1}{2\ (N-1)} \times \frac{sig(A_i, A_{i+1})}{\sum_{X \in \eta} sig(A_{i+1}, X)} + ... + \frac{1}{2\ (N-1)} \times \frac{sig(A_N, A_1)}{\sum_{X \in \eta} sig(A_N, X)} \\ & + ... + \frac{1}{2\ (N-1)} \times \frac{sig(A_{i-2}, A_{i-1})}{\sum_{X \in \eta} sig(X, A_{i-1})} \end{aligned} \tag{2}$$

Where $\eta$ is the node-set of the process model, N is the size of the node-set $\eta$, $sig:\eta\times\eta\to R_0^+$ is the priority relationship assigned to each pair of nodes $A_i, A_{i+1}\in\eta$, $rel_i$ is the relative importance of the N-element chain relationship, $A_i$ as the starting point and $A_{i\text{-}1}$ as the ending node. In addition, $A_0$ and $A_N$ are equivalent.

Similar to the binary conflict, the situations in which the N-ary conflict relationship is generated can also be divided into the following three categories: 1) N-ary cycle: N activities $\{A_1,A_2,...A_N\}$ form a cycle, that is after $\{A_1,A_2,...A_N\}$ is executed in sequence, $A_N$ can return to activity $A_1$ and start again. In this case, the priority relationship between these activities is allowed in the actual process and therefore needs to be preserved.

2) Exception: The process is executed in sequence $A_1\to A_2\to A_3,...,A_{N-1}\to A_N$, but there will be exceptions to $A_N\to A_1$ in the actual execution process. In this case, remove the exception edge in the weaker chain structure.

3) Concurrency: There is a parallel structure in N activities $\{A_1,A_2,...A_N\}$, and the log records the possible execution order. In this case, it is necessary to delete this conflicting sorting relationship. For example, if $A_1\to A_2$ and $A_3$ are parallel structures, that is, $A_1\to A_2$ and $A_3$ can occur in any order, the log will record the possible occurrences of $A_1\to A_2\to A_3$ and $A_3\to A_1\to A_2$. In this case, you need to delete $A_3\to A_1$ and $A_2\to A_3$ that cause conflicts. Relationship.

It can be seen from the above that the flight guarantee process model essentially does not have the possibility of N-ary cycles, so the causes of N-ary conflicts are exceptions and concurrent situations. The solutions are as follows.

Determine the offset between each chain relationship and the chain relationship with the greatest relative importance, as shown in formula (3).

$$S_m(A_1,A_2,...,A_N)=|\max \mathrm{rel}_i\text{-}\mathrm{rel}_m| \quad \mathrm{m}\in N \tag{3}$$

On this basis, determine the chain structure with the largest offset value, $\max S_m$, and find the edge with the least relative importance on this chain structure and delete it, namely $\min rel(A_i,A_{i+1})$.

The calculation of $\min rel(A_i,A_{i+1})$ is shown in formula (1). If the offset values of these chain structures are similar, it means that there is a less important parallel structure between the chains. This paper removes the edges that are different between the chain structures and the corresponding edges in the N-ary conflict relationship.

Unlike binary conflicts and N-ary conflicts, there is no concurrency in unary conflicts, but only self-circulation or exceptions. This situation can be resolved by creating a virtual node, removing the unary conflict, and introducing it into the fuzzy mining algorithm for edge filtering. As shown in Figure 3. After the loop is released, the priority relationship of virtual nodes B_1 and B→B_1 can be obtained.

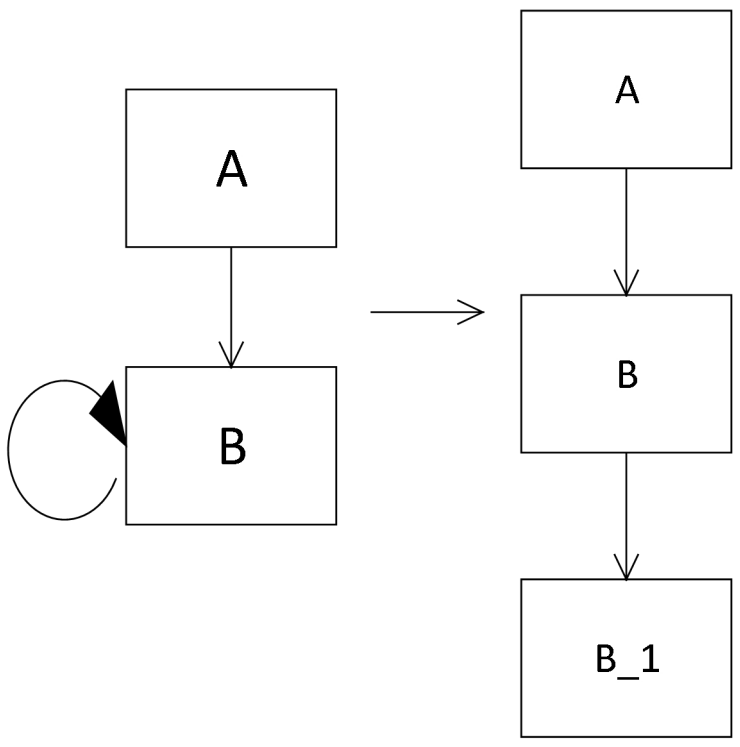

Figure 3: Conflict Resolution in Unary Relationships

## 3.2 Local causal structure construction

The second part of the framework focuses on building the MAG MB of the target variable based on the SLL algorithm under the assumption of relaxing causality sufficiency. Different from the MB of DAG, MB of MAG includes the area set and the node-set related to the area set in addition to the parent-child set and spouse set. The area set is defined as follows:

Definition 7 (District Set [7]): The district set of a target variable T in an MAG, denoted as dis(T),is a set of variables in which $\forall V_i \in dis(T)$ ,and the path from $V_i$ to T only contains bidirectional edges.

On this basis, the article [7] proposed a method for determining whether a variable belongs to the target variable district set. Suppose $V_j \in adj(T)$ , $V_i \in adj(V_j)$ , $V_k \in adj(V_i)$ ,if two triples $\{T,V_j,V_i\}$ , $\{T,V_j,V_i\}$ satisfy: $V_i \perp T|sepset(T,V_i)$ , $V_i \not\perp T|\{sepset(T,V_i) \cup V_j\}$ , $V_k \perp V_j|sepset(V_j,V_k)$ , $V_k \not\perp V_j|\{sepset(V_j,V_k) \cup V_i\}$, then $V_i \in dis(V_j)$.

The above theorem shows that the variable in sp(T) is a candidate variable for the district set variable in adj(T). From this, we can further conclude whether there is a bidirectional edge between the target variable T and the variable V, as shown below.

Corollary 8: If the triple $\{T,V_j,V_i\}$ satisfies $V_i \perp T|sepset(T,V_i)$, $V_i \not\perp T|\{sepset(T,V_i) \cup V_j\}$ and $adj(T)/V_j$ contain variable $V_m$ such that triple $\{V_m,T,V_j\}$ satisfies $V_j \perp V_m|sepset(V_m,V_j)$ , $V_j \perp V_m|\{sepset(V_m,V_j) \cup T\}$ , then $V_j$ and target variable T have bidirectional edges, that is, if there is $V_m \in adj(T)/V_j$ such that $V_m \in sp(V_j)$ in the neighborhood set of target variable T in v-structure $\{T,V_j,V_i\}$, then $V_j \in dis(T)$.

Through the above deduction, the district set adjacent to the target variable T can be judged. MAG MB includes the parents pa(T) of T, the children ch(T) of T, the spouses sp(T) of T, and the district set dis(T) of T, union of parents of each variable $V_i$ with the district set of T, that is, $U_{i=1}^{|dis(T)|}pa(V_i)$ ,denoted as pa(dis(T)), union of the district set each child $V_i$ of T, i.e., $U_{i=1}^{|ch(T)|}dis(V_i)$ , denoted as dis(ch(T)), union of parents of each variable $V_i$ within dis(ch(T)), that is, $U_{i=1}^{|U_{j=1}^{ch(T)}dis(V_i)|}pa(V_i)$.

Compared with the MB discovery algorithm that uses the independence test to find the target variable T, the score-based MB discovery algorithm relies on certain scoring criteria to learn the most suitable network structure for the data sample. It has the following characteristics:

Definition 9 (Local Score Consistency [28]): Let D be a set of data consisting of i.i.d samples from some distribution P.Let G be any BN structure and $G^{'}$ be the same structure as G. but with an edge from a node T to a node X. Let $Pa_X^G$ be the parent set of X in G. A score criterion s is locally consistent if, as the size of the Data D goes to infinity, the following two properties hold true:

(1) If $X \perp T \mid Pa_X^G$, then $s(G,D) < s(G^{'},D)$.

(2) If $X \perp T \mid Pa_X^G$, then $s(G,D) > s(G^{'},D)$.

Intuitively speaking, adding an arc can eliminate independent constraints that do not exist in the data generation distribution, thereby increasing the score. Adding an arc cannot eliminate such constraints and reduce the score. Therefore, the scoring function can replace constraints to construct a causal structure to some extent. For MAG structure learning, the existing DAG scoring function cannot be directly applied to MAG. $M^3C$ [29] and GSMAG [22] algorithms proposed a scoring function suitable for MAG, based on residual iterative conditional fitting to obtain the maximum likelihood estimation of a given MAG parameter. However, the M3HC and GSMAG algorithms make new assumptions on the data, which are not general. From Corollary 7, we can get the method for judging bidirectional edges. The core of the idea is to judge whether the spouse node of the target node T and the spouse node of the neighboring node V_j of T are in the other party's neighboring node. Therefore, this paper proposes a topology-based method to find the target node's neighborhood set and spouse set using the SLL[30] method. On this basis, use Corollary 7 to determine the bidirectional edge of the target node and the neighboring nodes. In this way, the neighboring district nodes of the target node are found, and the complete district set of the target variable is obtained through search the neighboring district nodes of the district nodes. The algorithm is shown below:

**Algorithm1 The SMMB Algorithm**

```
Input: Data D on node set N, a target node t ∈ N.
Output: MB(t)
    H*(t), ch(t), pa(t)=FINDNEIGHBORS( D, t)
    dis1,dis2,MB(t),dis=∅
    MB(t)= MB(t) ∪ H*(t)
    For q ∈ ch(t) ∪ t do
        S*(q)=FINDSPOUSES(D , q, H*(q), H*(v))
        dis*(q)=FINDDIS(D, q)
        dis2= dis ∪ dis*(q)
        numbers=|dis2 |
        if numbers>0 then
            While numbers don't change do
                for n∈ dis do
                    dis1=dis2
                    dis*(n)=FINDDIS(D ,n)
                    dis2= dis ∪ dis*(n)
                    numbers=|dis2|
                end for
        end if
        for m ∈ dis2 do
```

| | |
|---|---|
| **19:** | **$H^*(m), ch^*(m), pa^*(m)$=FINDNEIGHBORS( D, m)** |
| **20:** | **pa(dis)=pa(m)∪pa(dis)** |
| **21:** | **end for** |
| **22:** | **MB(t)= MB(t) ∪ dis ∪ pa(dis)** |
| **23:** | **end for** |

The first step of the SMMB algorithm is to search for the parent and child sets of the target node based on the SLL algorithm, and the fourth to fifth steps call the FINDSPOUSES of the SLL algorithm to find the spouse node of the target node and its child nodes. The idea is as follows: The FINDNEIGHBORS algorithm is divided into two stages. The first stage searches for the potential neighbor nodes of the target variable puts the nodes except the target node one by one into the set Z where only the target variable exists and calls the subroutine to learn Z. Put the learned potential neighbor nodes into Z to update Z to complete the search for the set of potential neighbors. The subroutine can use a dynamic programming algorithm or other precise algorithms. In this article, the commonly used precision algorithm, the GES [31] algorithm, is used as a subroutine, and its scoring function is as follows:

$$Score_{BIC} = \sum_{i=1}^{n} \sum_{j=1}^{q_i} \sum_{k=1}^{r_i} m_{ijk} \log_2 \frac{m_{ijk}}{m_{ij*}} - \sum_{i=1}^{n} \frac{q_i(r_i - 1)}{2} \log_2 m \tag{4}$$

The scoring form of the local structure formed by the target variable and its parent node is as follows:

$$BIC((X_i, \pi(X_i)) \mid D) = \sum_{j=1}^{q_i} \sum_{k=1}^{r_i} m_{ijk} \log_2 \frac{m_{ijk}}{m_{ij*}} - \frac{q_i(r_i - 1)}{2} \log_2 m \tag{5}$$

$$Score_{BIC} = \sum_{i=1}^{n} BIC((X_i, \pi(X_i)) \mid D) \tag{6}$$

Where m represents the number of samples, $m_{ijk}$ represents the number of samples that meet $X_i = k$ , $\pi(X_i)$=j in the data D, $X_i$ is the node variable in the network, and $\pi(X_i)$ represents the set of parent variables of node $X_i$. The node variable $X_i$ can take a discrete value or a continuous value, $k = 1, \dots, r_i$ , $r_i$ represents the state value of the child node, $j = 1, \dots, q_i$, $q_i$ represents the state value of the parent node, $m_{ij*} = \sum_{k=1}^{r_i} m_{ijk}$ , the network node variable X and the data D corresponds to the set Z and $D_Z$ in the algorithm. The transformation of Z and $D_Z$ is iterative. Change the scoring function of GES and the set of variables it can search by updating Z and $D_Z$. In the second stage, after searching for potential neighbor nodes, the symmetry correction is performed through Lemma 9 to remove false-positive nodes in the potential neighbor nodes to obtain the target variable's real neighbor nodes. FINDSPOUSES is similar to FINDNEIGHBORS. It searches for the target variable's potential mate set by calling a subroutine and then finds the true mate set of the target variable by forcing symmetry constraints.

In the sixth step, the SMMB algorithm calls the algorithm FINDDIS to search for the target variable's neighboring area nodes. According to Corollary 1, to determine whether a node belongs to the target node's area set, first find the spouse set of this node and the spouse set of the target node. The algorithm FINDDIS first searches the neighborhood set and spouse set of the target variable in steps 1-4, traverses the neighborhood of the target node in steps 5-7 to search for the neighborhood node of its neighborhood node, and step 8-9 find the variables belonging to the spouse node of T in the neighboring nodes and determine the collision node n of the V structure formed by this, steps 10-13 find the spouse node of the colliding node and determine whether it is in the neighborhood of the target node. If it is, then node n belongs to the area node of the target node.

| Algorithm2 The FINDDIS Algorithm |
|---|
| Input: Data D on node set N, a target node t∈N. |
| Output: MB(t) |
| 1: $H^*(t)$, ch(t), pa(t)=FINDNEIGHBORS( D, t) |
| 2: $S^*(t)$=FINDSPOUSES(D, t, $H^*(t)$, $H^*(v)$) |
| 3: dis(t)= ∅ |
| 4: numbers1=\| $H^*(t)$ \| |
| 5: if numbers1 >= 2 then |
| 6:   for n ∈ $H^*(t)$ do |
| 7:     $H^*(n)$=FINDNEIGHBORS(D ,n) |
| 8:     for $m \in H^*(n) \setminus t$ do |
| 9:       if $m \in S^*(t)$ then |
| 10:         $S^*(n)$=FINDSPOUSES(D,n, $H^*(n)$, $H^*(v)$) |
| 11:         for $m1 \in S^*(n)$ do |
| 12:           if $m1 \in H^*(t) \setminus n$ then |
| 13:             $dis(t) = dis(t) \cup n$ |
| 14:           end if |
| 15:         end for |
| 16:       end if |
| 17:     end for |
| 18:   end for |
| 19: end if |
| 20: return dis(t) |

Steps 10-16, based on searching the parent and child sets of the target node, the spouse set of the target node, and the neighboring area nodes of the target node, the FINDDIS algorithm is continuously iterated to search the area set of the target variable and its subset. Steps 18-21 through SLL to search for the parent set of the target variable's area set and the parent set of the subset area set.

## 4. Experiment

In order to evaluate the quality of the method proposed in this paper, in section 4.1, this paper uses the benchmark Bayesian network test data set alarm data set to test the method proposed in section 3.2, and evaluates the method proposed in section 3.2 with the indicator F-measure, which proves that it is suitable for common RFCI The superiority of the algorithm and the M3B algorithm. n section 4.2, use the flight guarantee data of China Xining Airport in July 2018, use the actual flight guarantee data at the airport to generate the airport flight guarantee process model, calculate the operation duration of each link as the performance indicator of each link, and extract the direction of the edges between nodes in the process model. The SMMB algorithm is used to construct a local causal model of performance indicators and flight launch delays and adjust the direction of the one-way edge in the causal model according to the direction between the extracted process model nodes. Finally, the MMHC algorithm is used to construct a local causal model of flight delays as a benchmark model to compare and analyze with the causal model constructed by SMMB. The local causal model constructed by the MMHC algorithm will also be adjusted according to the direction of the edges between nodes in the process model.

4.1 Causal discovery algorithm testing

The experimental test data source is the ALARM network, which contains 37 nodes and 46 edges. The network is a sparse network, which is considered a default standard for measuring the causal network construction program's level, and many algorithms and various programs have verified this data,and There is a standard network structure for comparison and reference. In order to test the performance of the above algorithm, this article uses ALARM data to randomly generate three sets of data. The first set of data includes 5 data sets of 2500 data instances, the second set of data includes 5 data sets of 5000 data instances, and the third The group data includes 5 data sets of 10,000 data instances. Then hide some common causes in the generated data set, and treat these hidden variables as potential common causes. Specific steps are as follows:

1) Do not hide any variables as latent variables, and this paper mine the local causal network of variable VTUB.

2) The hidden variable INT is used as a latent variable, and this paper mine the local causal network of VTUB, where INT is the potential common cause of the variable SHNT, the variable VLNG, and the variable PRSS.

3) Hidden variables INT and PMB are used as latent variables, and this paper mine the local causal network of VTUB. In this network, INT is the potential common cause of variable SHNT, variable VLNG, and variable PRSS, and variable PMB is the potential common cause of variable PAP and variable SHNT.

Use the data set generated by the above steps to compare the SMMB algorithm with RFCI and M3B algorithm, respectively. The RFCI and M3B algorithms are both constraint-based MAG discovery algorithms. There are three types of conditional independence tests, $G^2$ test for discrete variables, $Fisher's$ Z test for continuous variables with linear relations with additive Gaussian errors, and kernel-based test for continuous variables with nonlinearity and non-Gaussian.

This article is the same as the literature [16]. Both RFCI and M3B algorithms are tested, and the significance level of the test is set to 0.05. The test index used is F-measure.

**F-measure**: F-measure combines two indicators of prediction accuracy and recall and is defined as:

$$F\text{-}measure = \frac{2 \times \mathrm{Pr}\,ecision}{\mathrm{Pr}\,ecision + \mathrm{Re}\,call} \tag{7}$$

**Precision**: The prediction accuracy rate refers to the percentage of correctly predicted MNC to the total number of predicted MNI. It is used to evaluate the number of false positives in the output of the algorithm.

$$\mathrm{Pr}\,ecision = \frac{MNC}{MNI} \tag{8}$$

**Recall**: The recall rate refers to the ratio of the number of correctly predicted MNC to the total number . Used to evaluate the exact number in the output of the algorithm.

$$\mathrm{Re}\,call = \frac{MNC}{MNC + MNF} \tag{9}$$

MNF is the number of test samples that are not correctly identified. The index guarantee of F-measure combines two indicators of accuracy rate and recall rate. For the accuracy rate, recall rate, and F1 value obtained by each algorithm, this article takes the average of each group of 5 data sets. The comparison between SMMB algorithm and RFCI and M3B algorithm in accuracy, recall rate and F1 value is shown in Table 1. Figures 4, 5, and 6 are in turn without latent variables, INT as latent variables, and INT and PMB as latent variables. As shown in Tables 1, 2, and 3, the accuracy, recall rate and F1 value vary with sample size.

Table 1: When there is no latent variable, SMMB algorithm compares with RFCI and M3B algorithm in accuracy, recall rate and F1 value.

| without latent variables(None) | precision | | |
|---|---|---|---|
| | 2500 samples | 5000 samples | 10000 samples |
| SMMB | 0.765 | 0.65 | 0.6842858 |
| RFCI | 0.4217364 | 0.222867 | 0.2336904 |
| M3B | 0.5733334 | 0.633333 | 0.63 |
| | recall | | |
| SMMB | 0.6333334 | 0.6533334 | 0.6333334 |
| RFCI | 0.770909 | 1 | 1 |
| M3B | 0.3 | 0.333333 | 0.3666666 |
| | F1 | | |
| SMMB | 0.6761904 | 0.607619 | 0.6482718 |
| RFCI | 0.4199646 | 0.3586522 | 0.3718414 |
| M3B | 0.3919192 | 0.435556 | 0.459394 |

Table 2: When INT is used as a latent variable, SMMB algorithm is compared with RFCI and M3B algorithm in accuracy, recall rate and F1 value

| INT as latent variables(INT) | precision | | |
|---|---|---|---|
| | 2500 samples | 5000 samples | 10000 samples |
| SMMB | 0.692619 | 0.6166666 | 0.634286 |
| RFCI | 0.2378153 | 0.2260802 | 0.225484 |
| M3B | 0.6333334 | 0.6333334 | 0.625 |
| | recall | | |
| SMMB | 0.5428572 | 0.4571428 | 0.485714 |
| RFCI | 1 | 1 | 1 |
| M3B | 0.2857143 | 0.2857142 | 0.342857 |
| | F1 | | |
| SMMB | 0.590696 | 0.5111422 | 0.539134 |
| RFCI | 0.3835748 | 0.3687198 | 0.367719 |
| M3B | 0.3927272 | 0.3927272 | 0.429091 |

Table 3: When INT and PMB are used as latent variables, SMMB algorithm is compared with RFCI and M3B algorithm in accuracy, recall rate and F1 value

| INT and PMB as latent variables(INT\PMB) | precision | | |
|---|---|---|---|
| | 2500 samples | 5000 samples | 10000 samples |
| SMMB | 0.6157738 | 0.5833334 | 0.6342858 |
| RFCI | 0.2232481 | 0.223421 | 0.219775 |
| M3B | 0.6333334 | 0.6333334 | 0.6071428 |
| | recall | | |
| SMMB | 0.5714285 | 0.4571428 | 0.4857142 |
| RFCI | 0.9642858 | 1 | 0.9714286 |
| M3B | 0.2857143 | 0.2857142 | 0.3142858 |
| | F1 | | |
| SMMB | 0.58837 | 0.494732 | 0.5391342 |
| RFCI | 0.3625138 | 0.3651232 | 0.3583038 |
| M3B | 0.3927272 | 0.3927272 | 0.4062338 |

From the accuracy comparison chart, we can clearly see that the SMMB algorithm and the M3B algorithm are relatively close in accuracy, and both are better than the RFCI algorithm. In the comparison of the recall rate, the RFCI algorithm is much higher than the SMMB algorithm and the M3B algorithm.It can be seen that the RFCI algorithm contains more redundant nodes than the causal network discovered by the M3B and SMMB algorithms. Compared with the M3B algorithm, the SMMB algorithm has a certain degree of competition in accuracy, but the SMMB algorithm has a higher recall rate than the M3B algorithm.As a result, in terms of comprehensive evaluation index F1, the F1 value of the SMMB algorithm is better than that of the M3B algorithm and the RFCI algorithm.

SMMB algorithm, RFCI algorithm, and M3B algorithm are algorithms for constructing causal networks based on topology, and all need to find adjacent variables of a given target variable first.For the RFCI algorithm, the key is to find the correct graph skeleton from the data set. For the SMMB algorithm and M3B algorithm, the key is to find the neighboring nodes of the target variable. The RFCI algorithm uses the PC-stable algorithm to find the network skeleton, while the M3B algorithm uses the AdjV algorithm to find the neighboring nodes of the target variable. SMMB algorithm uses the score-based SLL algorithm framework when looking for the target node's parent-child set and spouse set. Compared with the constraint-based method such as the AdjV algorithm and the PC-stable algorithm, this algorithm searches the target node's neighborhood set. It has more advantages.

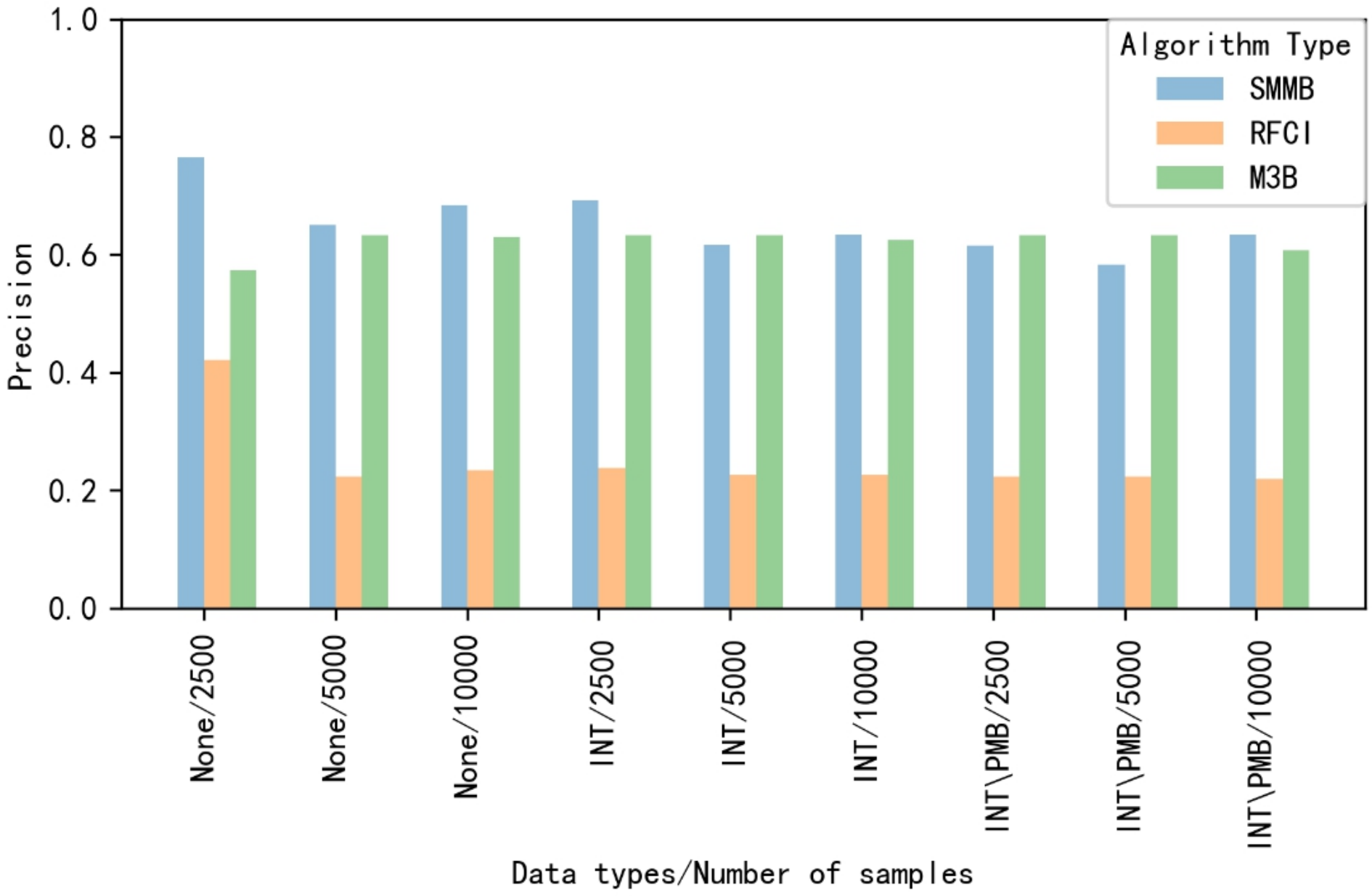

Figure 4: Comparison of SMMB, RFCI, and M3B in precision

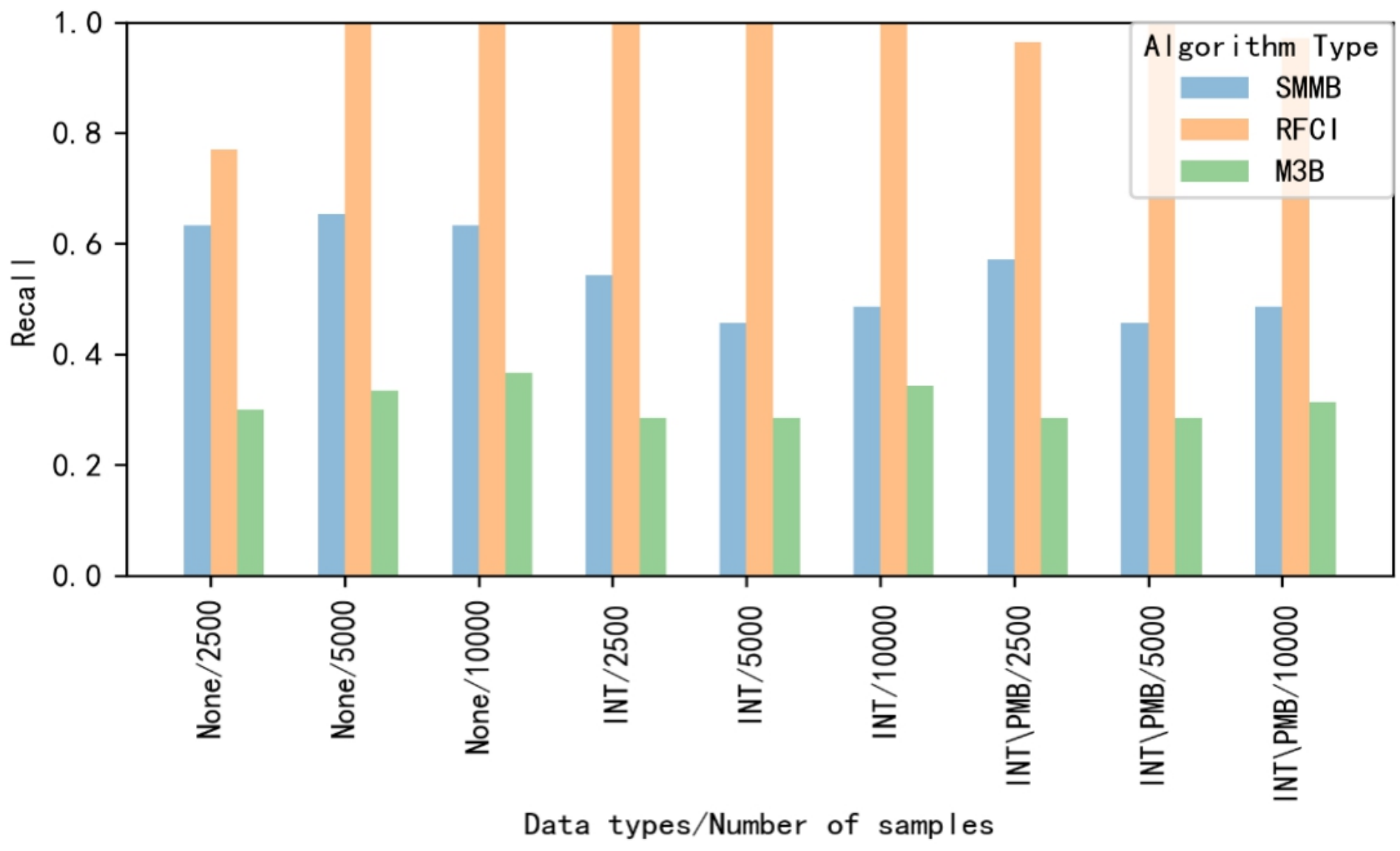

Figure 5: Comparison of SMMB, RFCI, and M3B in recall

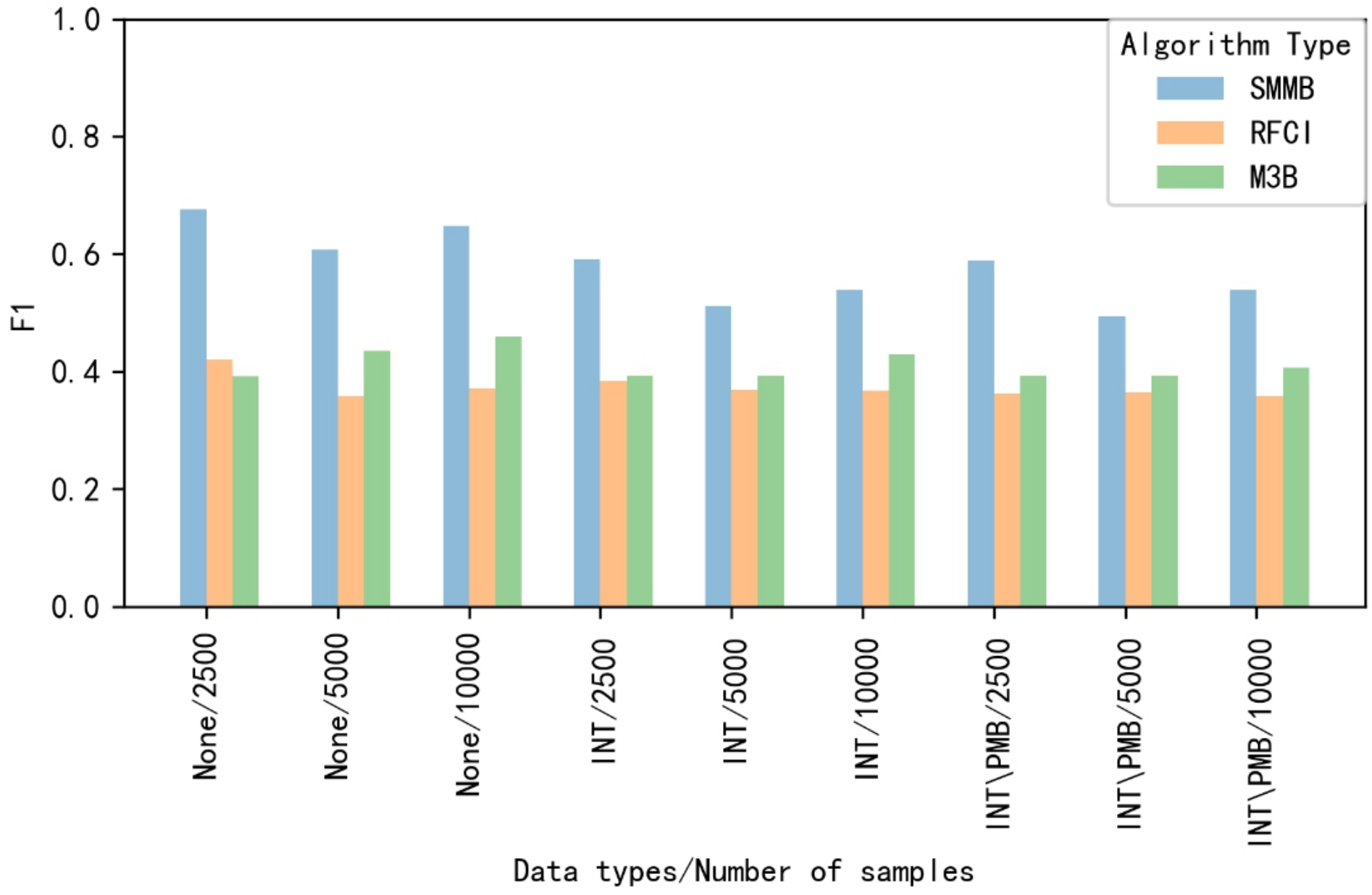

Figure 6. Comparison of SMMB, RFCI, and M3B in F1

## 4.2 Cause and effect of flight service process

The ProM used in this paper is used as a tool for mining the flight service process, using the flight guarantee data of Xining Airport in China in July 2018 for a case study. There are 122839 pieces of this data, which record the case, type, Activity, Resource, Timestamp, and other information of Xining Airport from July 1st to July 31st. The data sample is shown in Table 4 below.

Table 4: Example log of flight service event t Xining Airport in July 2018

| Case | type | Activity | Resource | Timestamp | lifecycle:transition |
|---|---|---|---|---|---|
| A | task | DROPOFFEND | a | 2018-07-01T11:54:27.000 | complete |
| A | task | DROPOFFCHECK | a | 2018-07-01T11:55:19.000 | complete |
| A | task | LUGLOADPERSONARR | a | 2018-07-01T11:55:26.000 | complete |

The fuzzy process mining algorithm plug-in in the ProM tool performs process mining on the flight service event log. The retention threshold and the ratio threshold are set to 0.27 and 0.35, respectively, to obtained the flight service process model. As mentioned in section 3.1, The process model is optimized for the N-ary conflict and the unary conflict resolution. The optimized model is shown in Figure 7. For the node's evaluation index, this paper calculates the timestamp difference between the previous node and the next node as the duration of the previous node to evaluate the service operation efficiency of each node. For the link structure with parallel relationship, according to the link structure's duration is used as the basis for competition, and the longer part is selected as the overall duration of the parallel structure. In addition, the departure time of the node's front station and the node's own station is used as the evaluation index. By looking for the causal relationship between the flight launch delay time and the evaluation indicators of each flight guarantee node, we can analyze which nodes of the flight guarantee cause the final delay. Import the obtained evaluation index of each flight node into the SMMB algorithm, and the target node is flight delay. The resulting causal network of flight guarantee is shown in Figure 8 below. Figure 9 shows the local causal model of flight delays obtained after importing the evaluation indicators of each node of the flight guarantee into the PC algorithm. This is extracted from the complete causal model. The meaning of each node letter in the figure is shown in Table 4.

It can be seen from Figure 9 that CHECKINSTART_ALANDINTIME, REGEAR_REALTAKEOFF,TRACTORBIND_TRACTORSTART,MANIFESTCONVEY_CLOSECABINGATE,CLEANPERSONARR_CLEANSTART,REARCABINTRANSPORTCAREND_TRACTORBIND,DSHUTTLEARR_BOARDINGSTART,TAKEOFFATTHEFRONTSTATION_PUSHPERSONARR,PASSENGERUPPERSONARR_DSHUTTLEARR, DELIVERYPERSONARR_REGEAR are considered to be the main causes of flight delays. However, in Figure 8, there are bidirectional edges between CHECKINSTART_ALANDINTIME,REGEAR_REALTAKEOFF, TRACTORBIND_TRACTORSTART, MANIFESTCONVEY_CLOSECABINGATE, and flight delays, which means that there are unobservable latent variables that affect the two sides of the bidirectional edges connection. This is also the advantage of the MAG graph to represent the causal model. The MAG graph can represent the existence of latent variables through bidirectional edges, avoiding the influence of confounding effects caused by latent variables. The causal network diagram discovered by the MMHC algorithm does not consider the existence of latent variables, leading to the mistaken belief that CHECKINSTART_ALANDINTIME,REGEAR_REALTAKEOFF,TRACTORBIND_TRACTORSTART,and MANIFESTCONVEY_CLOSECABINGATE are the direct causes of flight delays. There is no causal relationship in nature. Therefore, improving the four links of CHECKINSTART_ALANDINTIME, REGEAR_REALTAKEOFF,TRACTORBIND_TRACTORSTART, and MANIFESTCONVEY_CLOSECABINGATE in airport flight guarantee cannot effectively improve airport flight delays. Figures 8 and 9 all believe that CLEANPERSONARR_CLEANS-TART,REARCABINTRANSPORTCAREND_TRACTORBIND,DSHUTTLEARR_BOARDINGSTART,TAKEOFFATTHEFRONTSTATION_PUSHPERSONARR,and PASSENGERUPPERSONARR_DSHUTTLEARR are the direct causes of flight delays. Therefore, if the airport wants to change the current situation of flight delays, it can start by improving the interval time of the five links of CLEANPERSONARR_CLEANSTART, REARCABINTRANSPORTCAREND_TRACTORBIND,DSHUTTLEARR_BOARDINGSTART,TAKEOFFATTHEFRONTSTATION_PUSHPERSONARR, and PASSENGERUPPERSONARR_DSHUTTLEARR, optimize the efficiency of cleaning personnel and release personnel, and reasonably plan the driving route of tractors and shuttles, to improve the status of flight delays at the airport.

## 5. Conclusions and Future Work

This paper uses a new framework to automatically discover the root cause of business process performance problems. Compared with the automatic causal discovery of business process performance based on the Granger causality test proposed by some researchers, the method proposed in this paper uses the SMMB algorithm. Instead of the Granger causality test to solve the influence of potential confounding effects, the node pairs with latent variables are marked by bidirectional edges. However, this article only discovers the causal relationship between business performance indicators represented by flight delays and node operation duration and does not explore the specific causal effects of each node duration on flight delays. This problem is essential that a causal effect estimation problem when covariates are missing. In future work, we will further explore this problem by estimating the causal effect of each node on flight delays and looking for nodes that have a greater impact on delays; take appropriate measures to these nodes to improve the business performance of the airport.

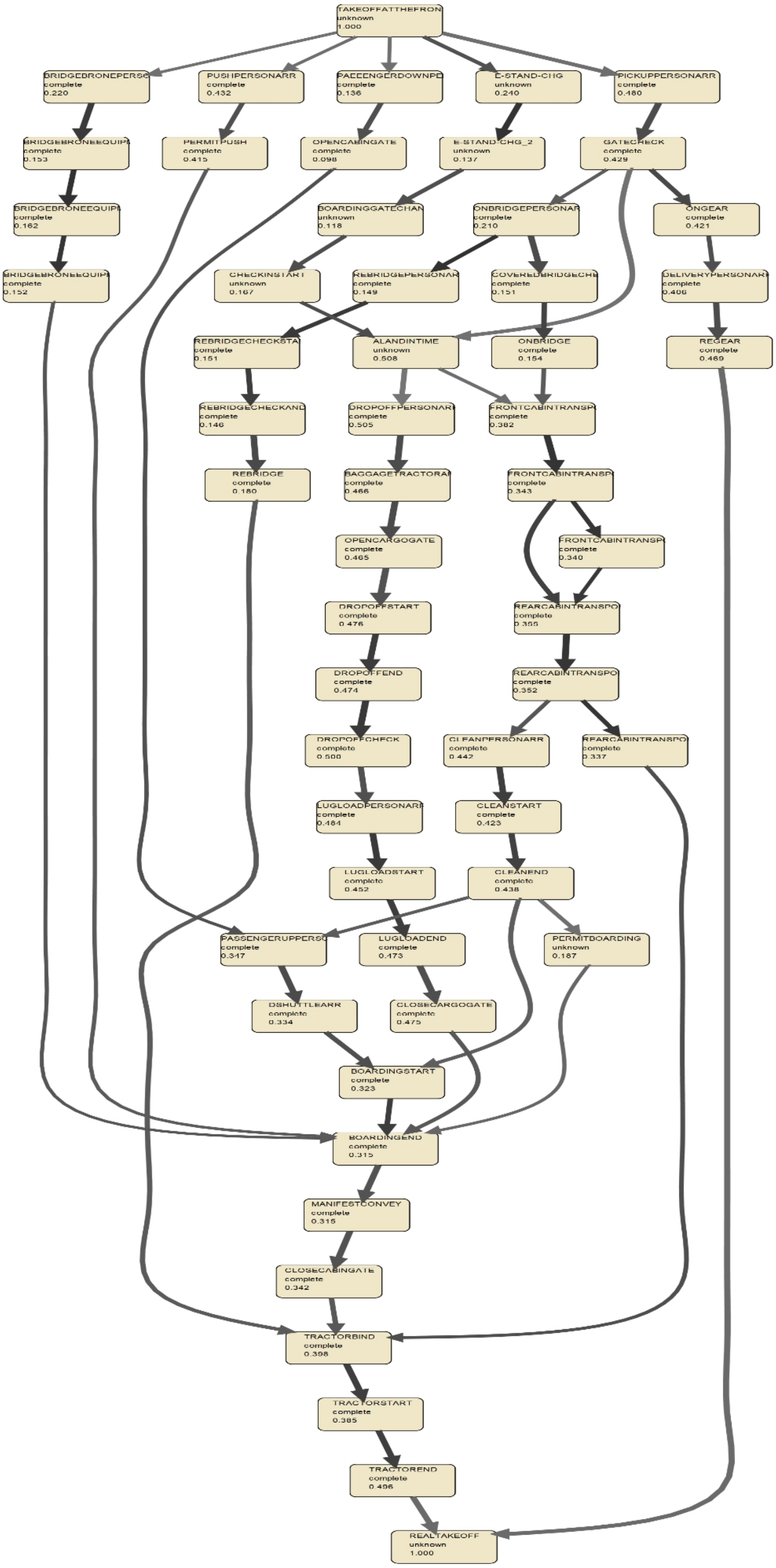

Figure 7: Process model of flight service

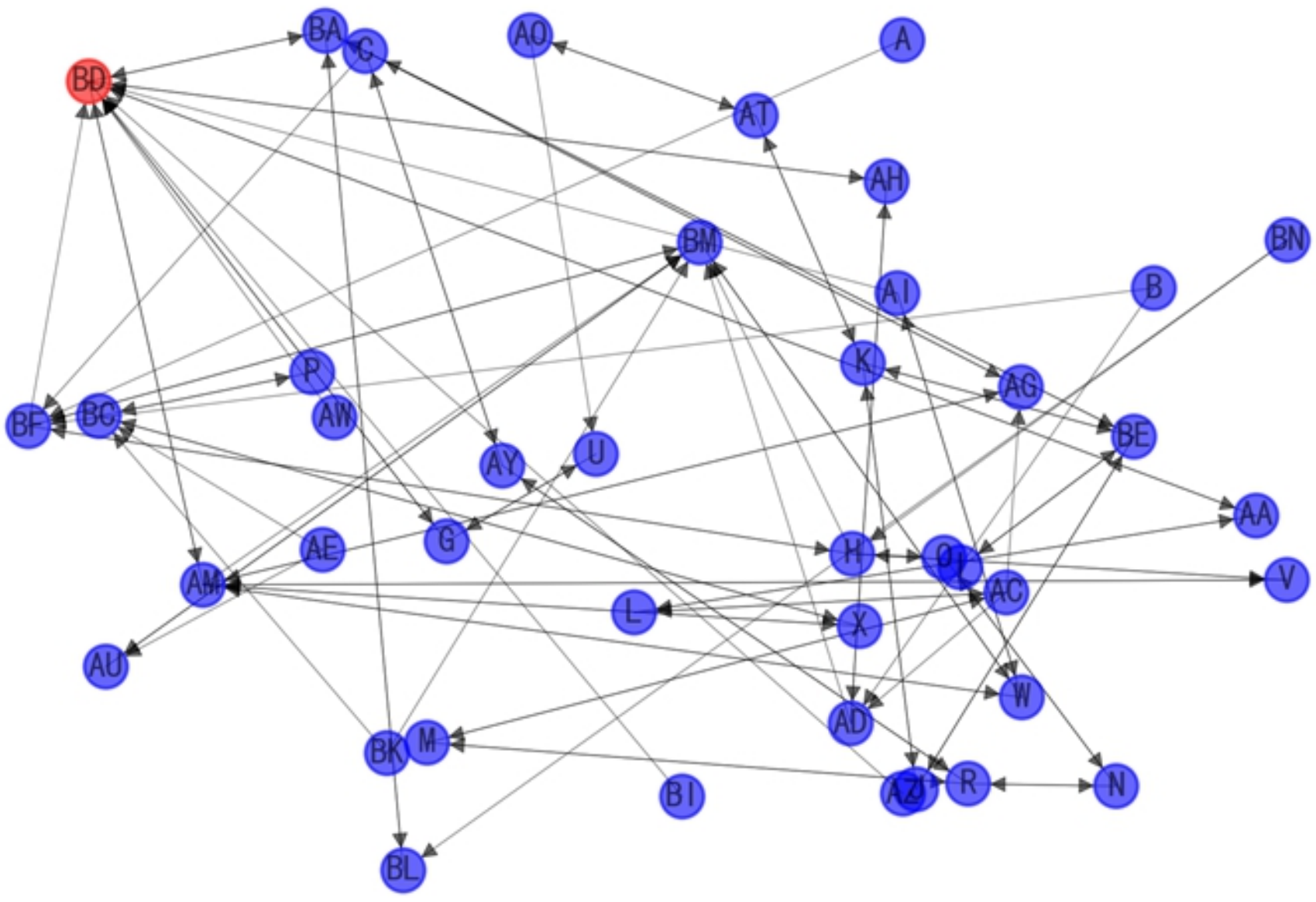

Figure 8: The cause and effect diagram of the partial process of SMMB flight guarantee

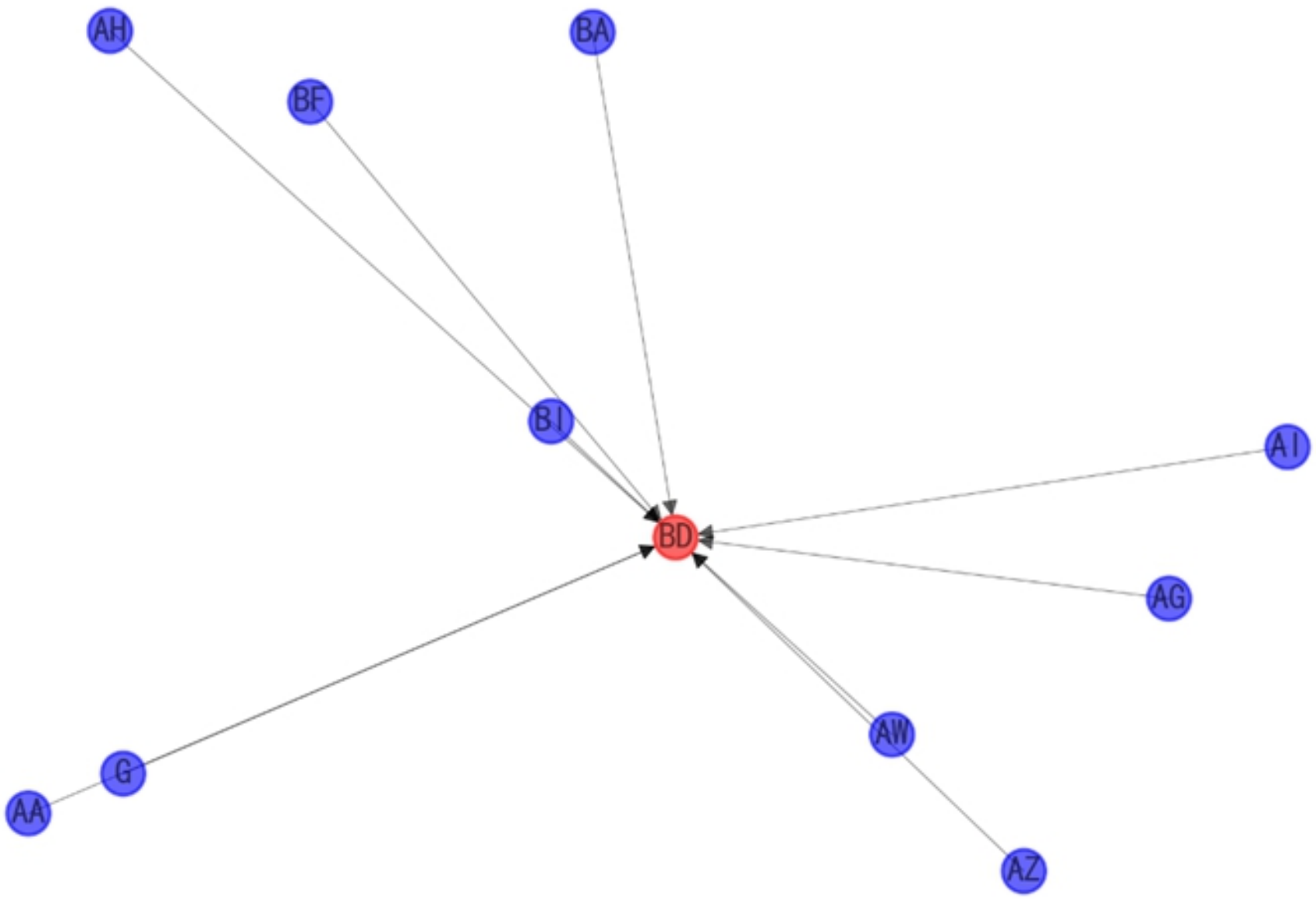

Figure 9: The causal graph of the local process of PC flight service

Table 5. Notes on the causal graph nodes of flight service

| A | B | C | G | H | I | J |
|---|---|---|---|---|---|---|
| TAKEOFFATTHEFRONTSTATION_E-STAND-CHG | TAKEOFFATTHEFRONTSTATION_PICKUPPERSONARR | TAKEOFFATTHEFRONTSTATION_PAEEENGERDOWNPER | CHECKINSTART_ALANDINTIME | ALANDINTIME_DROPOFFPERSONARR | BRIDGEBRONEPERSONARR_BRIDGEBRONEEQUIPMEN | BRIDGEBRONEEQUIPMENTARR_BRIDGEBRONEEQUIPME |
| | | SONARR | | | TARR | NTSTART |
| K | L | M | N | O | P | R |
| BRIDGEBRONEEQUIPMENTSTART-BRIDGEBRONEEQUIPMENTEND | DROPOFFPERSONARR_BAGGAGETRACTORARR | BAGGAGETRACTORARR_OPENCARGOGATE | OPENCARGOGATE_DROPOFFSTART | DROPOFFSTART_DROPOFFEND | DROPOFFEND_DROPOFFCHECK | LUGLOADPERSONARR_LUGLOADSTART |
| U | V | W | X | AA | AC | AD |
| PUSHPERSONARR-PERMITPUSH | PERMITBOARDING_BOARDINGEND | CLOSECARGOGATE_BOARDINGEND | BRIDGEBRONEEQUIPMENTEND-BOARDINGEND | MANIFESTCONVEY_CLOSECABINGATE | GATECHECK_ONGEAR | GATECHECK_ALANDINTIME |
| AE | AG | AH | AI | AM | AO | AT |
| GATECHECK-ONBRIDGEPERSONARR | DELIVERYPERSONARR_REGEAR | REGEAR_REALTAKEOFF | CLEANPERSONARR_CLEANSTART | BOARDINGSTART_BOARDINGEND | ONBRIDGEPERSONARR_COVEREDBRIDGECHECK | COVEREDBRIDGECHECK_ONBRIDGE |
| AU | AW | AY | AZ | BA | BC | BD |
| FRONTCABINTRANSPORTCARSTART-FRONTCABINTRANSPORTCAREND | REARCABINTRANSPORTCAREND_TRACTORBIND | PAEEENGERDOWNPERSONARR_OPENCABINGATE | DSHUTTLEARR_BOARDINGSTART | TRACTORBIND_TRACTORSTART | TRACTOREND_REALTAKEOFF | FLIGHT DELAY |
| BE | BF | BI | BK | BL | BM | BN |
| TAKEOFFATTHEFRONTSTATION-BRIDGEBRONEP | TAKEOFFATTHEFRONTSTATION_PUSHPERSONA | PASSENGERUPPERSONARR_DSHUTTLEARR | FRONTCABINTRANSPORTCARSTART_REARCABI | ONBRIDGE_FRONTCABINTRANSPORTCARBIND | PERMITPUSH_BOARDINGEND | ALANDINTIME_FRONTCABINTRANSPORTCARBI |
| ERSONA RR | RR | | NTRANS PORTCA RBIND | | | ND |

## 6. Data Availability

The airport ground support database was purchased from Airports Council International (ACI). To have access, the researcher must get in touch with this organization and acquire the data. The alarm database comes from the bnlearn package.

## 7. Conflicts of Interest

The authors declare that they have no conflicts of interest.

## 8. Acknowledgments

This research was supported by Key R&D Program of Sichuan Province (under Gaint no.2020YFG0050 and no.2020YFG0058) and the Sichuan Youth Science and Technology Innovation Team Special Program (under Gaint no.2019JDTD 0001). These program are supported by the Sichuan Science and Technology Program.

## 9. References:

[1] Civil Aviation Administration of China. “2019 Statistical Bulletin on the Development of the Civil Aviation Industry.” Civil Aviation Management, pp. 46-47, 2020.

[2] Xing Zhiwei, WEI Zhiqiang, LUO Qian. “Flight support service process modeling method based on colored time Petri net,” Journal of Systems Engineering and Electronics, vol.40, no.464, pp.109-114, 2018.

[3] HAO Jing-qi,YANG Wen-dong,TANG Xiao-wei. “Simulation for Platform Truck Resource Allocation in Hub Airport Apron.” Journal of the Wuhan University of Technology, vol.35, no.12, pp.85-91, 2013.

[4] Xing Zhiwei，Tang Yunxiao. “Flight Support Service Time Estimation of Hub Airport .” JOURNALOFSYSTEMSIMULATION, vol.29, no.11, pp.2856-2864, 2017.

[5] Kovynyov I , Mikut R. “Digital technologies in airport ground operations.” NETNOMICS Economic Research and Electronic Networking, vol.20, no.3, pp.1-30, 2019.

[6] Rudnitckaia, Julia. “Process Mining. Data science in action.” University of Technology, Faculty of Information Technology, pp.1-11, 2015.

[7] Yu K, Liu L, Li J, et al. “Mining Markov blankets without causal sufficiency.” IEEE transactions on neural networks and learning systems, vol.29, no.12, pp.6333-6347, 2018.

[8] Ciccio C D , Marrella A , Russo A . “Knowledge-Intensive Processes: Characteristics, Requirements and Analysis of Contemporary Approaches.” Journal on Data Semantics, vol.38, no.3, pp.29-57, 2015.

[9] Jiao-jiao Wang. “Research on Some Key Issues of Business Process Management Based on Event Log,” HANGZHOU DIANZI UNIVERSITY, 2015.

[10] Corallo A , Lazoi M , Striani F . “Process mining and industrial applications: A systematic literature review.” Knowledge and Process Management, vol.40, no.3, pp. 225-233, 2020.

[11] Van D , Weijters T , Maruster L . “Workflow mining: discovering process models from event logs.” IEEE Transactions on Knowledge & Data Engineering, vol.16, no.9, pp.1128-1142, 2004.

[12] Weijters, A. J. M. M., and J. T. S. Ribeiro. “Flexible heuristics miner (FHM).” 2011 IEEE symposium on computational intelligence and data mining (CIDM). IEEE, 2011.

[13] Montani S , Leonardi G , Quaglini S , et al. “A knowledge-intensive approach to process similarity calculation.” Expert Systems with Applications, vol.42, no.9, pp.4207-4215, 2015.

[14] Solé, Marc, and Josep Carmona. "Amending C-net discovery algorithms." Proceedings of the 28th Annual ACM Symposium on Applied Computing. 2013.

[15] CW Günther, Aalst W. “Fuzzy Mining – Adaptive Process Simplification Based on Multi-perspective Metrics.” Business Process Management, International Conference, Bpm, Brisbane, Australia, September. Springer-Verlag, 2007.

[16] Spirtes P, Glymour C, Scheines R. “Causation, Prediction, and Search,” 2nd Edition. The MIT Press, 2000.

[17] Pearl J. “Probabilistic Reasoning in Intelligent Systems: Networks of Plausible Inference.” Artificial Intelligence, vol.48, no.8, pp.117-124, 1990.

[18] Tsamardinos I, Aliferis C F, Statnikov A R, et al. “Algorithms for large scale Markov blanket discovery[C]//FLAIRS conference, vol.2, pp.376-380, 2003.

[19] Tsamardinos, Ioannis, Constantin F. Aliferis, and Alexander Statnikov. “Time and sample efficient discovery of Markov blankets and direct causal relations.” Proceedings of the ninth ACM SIGKDD international conference on Knowledge discovery and data mining, 2003.

[20] Aliferis, Constantin F., et al. “Local causal and Markov blanket induction for causal discovery and feature selection for classification part I: algorithms and empirical evaluation.” Journal of Machine Learning Research, 2010.

[21] Aliferis, C. F., Statnikov, A., Tsamardinos, I et al. “Local causal and Markov blanket induction for causal discovery and feature selection for classification part II: analysis and extensions.” *Journal of Machine Learning Research*, 2010.

[22] Triantafillou, S., & Tsamardinos, I. “Score-based vs Constraint-based Causal Learning in the Presence of Confounders.” In *CFA@ UAI*, pp. 59-67, 2016.

[23] Richardson, T., & Spirtes, P. “Ancestral graph Markov models.” *The Annals of Statistics*, vol.30, no.4, pp.962-1030, 2002.

[24] Van der Aalst, W., Adriansyah, A., & van Dongen, B. “Replaying history on process models for conformance checking and performance analysis.” *Wiley Interdisciplinary Reviews: Data Mining and Knowledge Discovery*, vol.2, no.2, pp.182-192, 2012.

[25] González, L. S., Rubio, F. G., González, F. R., & Velthuis, M. P. (2010). “Measurement in business processes: a systematic review.” *Business Process Management Journal, vol.16, no.1, pp.1463-7154, 2010.*

[26] Hompes, B. F., Maaradji, A., La Rosa, M., Dumas et al, “Discovering causal factors explaining business process performance variation.” In *International Conference on Advanced Information Systems Engineering*, vol.10253, pp.177-192, 2017.

[27] Günther, C. W., & Van Der Aalst, W. M. “Fuzzy mining–adaptive process simplification based on multi-perspective metrics.” In *International conference on business process management*. Springer, Berlin, Heidelberg. vol.4714, pp.328-343, 2007.

[28] Chickering, D. M. (2002). “Optimal structure identification with greedy search. *Journal of machine learning research*,” *vol.3*, pp.507-554, 2002.

[29] Tsirlis, K., Lagani, V., Triantafillou, S., & Tsamardinos, I. “On scoring maximal ancestral graphs with the max–min hill climbing algorithm.”*International Journal of Approximate Reasoning*, vol.102, no.7, pp.74-85, 2018.

[30] Niinimaki, T., & Parviainen, P. "Local structure discovery in Bayesian networks." *arXiv preprint arXiv:1210.4888*, 2012.
[31] Silander, T., & Myllymaki, P. "A simple approach for finding the globally optimal Bayesian network structure". *arXiv preprint arXiv:1206.6875*, 2012.